\documentclass[final,authoryear,5p,times,twocolumn]{elsarticle}

\usepackage{hyperref}

\journal{Medical Image analysis}

\usepackage[utf8]{inputenc}
\usepackage{amsmath}
\usepackage{amsfonts}
\usepackage{amssymb}
\usepackage{tabularx,multirow}
\usepackage{float}
\usepackage[lofdepth,lotdepth]{subfig}
\usepackage{rotating}

\biboptions{authoryear}

\begin{document}

\begin{frontmatter}

\title{Prediction of final infarct volume from native CT perfusion and treatment parameters using deep learning}

\author[mirc,mic,ico]{David Robben \corref{cor1} }
\ead{david.robben@kuleuven.be}
\author[amsterdam]{Anna M.M. Boers}
\author[amsterdam]{Henk A. Marquering}
\author[antonius]{Lucianne L.C.M. Langezaal}
\author[amsterdam]{Yvo B.W.E.M. Roos}		
\author[maastricht]{Robert J. van Oostenbrugge}
\author[maastricht]{Wim H. van Zwam}
\author[erasmusneuro]{Diederik W.J. Dippel}
\author[amsterdamrad]{Charles B.L.M. Majoie}
\author[erasmusrad]{Aad van der Lugt}
\author[neuro,vib,uz]{ Robin Lemmens }
\author[mirc,mic]{ Paul Suetens }

\cortext[cor1]{Corresponding author}

\address[mirc]{Medical Imaging Research Center (MIRC), KU Leuven, Leuven, Belgium}
\address[mic]{Medical Image Computing (MIC), ESAT-PSI, Department of Electrical Engineering, KU Leuven, Leuven, Belgium}
\address[ico]{Icometrix, Leuven, Belgium}

\address[amsterdam]{Amsterdam UMC, Amsterdam, the Netherlands}
\address[amsterdamrad]{Amsterdam UMC, location AMC, Amsterdam, the Netherlands; Department of Radiology and Nuclear Medicine}
\address[maastricht]{Maastricht UMC, Maastricht, the Netherlands}
\address[erasmusneuro]{Department of Neurology, Erasmus MC, University Medical Center Rotterdam, the Netherlands}
\address[erasmusrad]{Department of Radiology  \& Nuclear Medicine, Erasmus MC, University Medical Center Rotterdam, the Netherlands}
\address[antonius]{St. Antonius Ziekenhuis, Nieuwegein, the Netherlands}

\address[neuro]{Department of Neurosciences, Experimental Neurology, and Leuven Brain Institute (LBI), KU Leuven –  University of Leuven, Leuven, Belgium }
\address[vib]{VIB, Center for Brain \& Disease Research, Laboratory of Neurobiology, Leuven, Belgium}
\address[uz]{University Hospitals Leuven, Department of Neurology, Leuven, Belgium }

\begin{abstract}
CT Perfusion (CTP) imaging has gained importance in the diagnosis of acute stroke.
Conventional perfusion analysis performs a deconvolution of the measurements and thresholds the perfusion parameters to determine the tissue status.
We pursue a data-driven and deconvolution-free approach, where a deep neural network learns to predict the final infarct volume directly from the native CTP images and metadata such as the time parameters and treatment.
This would allow clinicians to simulate various treatments and gain insight into predicted tissue status over time.
We demonstrate on a multicenter dataset that our approach is able to predict the final infarct and effectively uses the metadata.
An ablation study shows that using the native CTP measurements instead of the deconvolved measurements improves the prediction.
\end{abstract}


\end{frontmatter}


\section{Introduction}
Ischemic stroke, a major cause of mortality and disability worldwide, is an acute disease where the blood supply to the brain is hindered due to the occlusion of an artery.
Due to the reduction in perfusion, neuronal functioning is impaired and
if perfusion is not re-established, brain tissue becomes irreversibly damaged. 
Acute treatment aims at reopening the blocked artery through treatment with intra-venous thrombolytics and/or mechanical thrombectomy.
Since both treatment options entail considerable costs and side effects (e.g. increased risk of haemorrhage), selection of patients who might benefit is important.
In early studies, patient selection was mainly based on the time since stroke onset; e.g. intravenous thrombolysis with tPA is shown to be beneficial within 4.5 hours \citep{Hacke2008}.
Recent studies typically select patients based on both time parameters and advanced neuroimaging, allowing patient-specific assessment of the benefits and risk of treatment.

One commonly used imaging modality for acute stroke diagnosis is CT Perfusion (CTP), which consists of a series of 3D CT scans acquired after intra-venous injection of contrast agent.
The resulting 4D image shows the passage of contrast agent through the brain:
 in each spatial voxel we have a time-attenuation curve showing the variation of intensity due to the contrast agent.
Much research has gone into the quantification of these images, 
 particularly aiming at the estimation of perfusion parameters such as cerebral blood flow (CBF), cerebral blood volume (CBV) and Tmax.
The most used approach for perfusion analysis works as follows \citep{Fieselmann2011}.
First the time-attenuation curves are converted to time-concentration curves that show the concentration of contrast agent over time.
In CTP, there is a linear relationship between the change in attenuation and the concentration.
Then an arterial input function (AIF) is determined: this is the time-concentration curve in one of the large feeding arteries of the brain.
Subsequently, the time-concentration curve in each voxel is deconvolved with the AIF, resulting in deconvolved time-concentration curves.
These curves correspond to what would have been measured if the contrast bolus was infinitely short and infinitely concentrated in the feeding artery -- indeed, these curves are impulse response functions (IRF).
As such, the deconvolved time-concentration curves are no longer influenced by the contrast injection protocol or the cardiac system of the particular patient.
Finally, under some reasonable assumptions, the perfusion parameters can be derived from these curves.
For example
$\text{CBF}	\propto \max_t \text{IRF}(t)$ and $\text{Tmax} 	= \operatorname*{argmax}_t \text{IRF}(t)$.

The deconvolution operation plays a central role in current perfusion analysis.
However, deconvolution depends crucially on the accurate selection of an AIF.
Additionally, deconvolution is a mathematically ill-posed problem and that is problematic given the low signal to noise ratio of perfusion images \citep{Fieselmann2011}.
Perfusion analysis software accounts for this by preprocessing the images and by regularizing the deconvolution.
The preprocessing mainly aims at reducing the noise through motion correction, temporal and spatial smoothing, and possibly spatial downsampling.
The regularization of the deconvolution suppresses the high frequency signal in the reconstructed impulse response function.
This can be done in singular value decomposition (SVD) based deconvolution by regularizing the singular values, e.g. using Tikhonov regularization.
Nevertheless, the deconvolution-based perfusion parameters remain noise sensitive and research for improved algorithms \citep{Boutelier2012} or even deconvolution-free summary parameters \citep{Meijs2016} remains ongoing.

The resulting perfusion parameters are used to assess the tissue status of the brain where -- apart from healthy tissue -- two types are distinguished.
Tissue that is already irreversibly damaged is called the infarct core.
Tissue that is at risk, i.e. tissue that will undergo infarction if not eventually reperfused, is called the penumbra.
The combination of infarct core and penumbra is called the perfusion lesion.
In current clinical practice, the perfusion lesion and core are determined by thresholding the perfusion parameters.
The optimal choice of perfusion parameters and thresholds depends on the deconvolution method \citep{Bivard2013}.

Knowledge about the volumes of the core and the penumbra is of great clinical importance, as they are used to determine which treatment a patient should get \citep{Albers2018,Nogueira2018}.
However, the method that is used to analyze the CTP images plays an important role in the accuracy of these volume estimations \citep{Fahmi2012,Bivard2013}.
Although great progress has been made in the previous years, there is room and need for improvement.

The binary distinction between core and penumbra is somewhat artificial:
 as time passes, the core will become larger, as more and more penumbral tissue becomes irreversibly damaged.
The infarct growth rate differs between patients \citep{Wheeler2015,Guenego2018}.
It depends on factors such as the location of the thrombus and the patient's vascular connectivity:
 there is considerable inter-subject variation in the amount of vascular redundancy and hence in the amount collateral blood supply to the affected region \citep{Liebeskind2003}.
Knowledge about the growth rate of the core is of clinical importance,
 as it allows assessment of the relevance of transferring a patient to a comprehensive stroke center depending on transport time.
Hence it is valuable to not only predict core and penumbra from an acute CTP scan, but also how the core would evolve over time.
Additionally, not every mechanical thrombectomy procedure achieves complete reperfusion. 
Although a more complete recanalization is arguably always better \citep{Kleine2017}, it is interesting to take this into account in the model
 as it might give interventional radiologists information about the added value of an additional attempt after a partial recanalization. 

\subsection{Contributions}
In this paper we propose to train a deep neural network to predict the final infarct from acute CTP images.
The prediction takes into account not only the CTP measurements,
 but also the treatment parameters such as the time-to-recanalization and the completeness of recanalization.
Once the network is trained, it can make predictions for new patients based on hypothetical treatment schemes.
This allows to predict the ischemic core at baseline (i.e. the predicted final infarct in case of an immediate perfect recanalization), the perfusion lesion (i.e. the predicted final infarct in case of no recanalization)
 or any intermediate scenario.
The latter can be relevant if e.g. we first need to transfer the patient and achieve recanalization at a particular later time point.
This provides clinicians with additional information on the impact of potential treatment options.
We show that our method takes this information effectively into account resulting in better predictions.

Another contribution of this work is that the predictions are made directly from the native CTP images while no explicit deconvolution or calculation of perfusion parameters is performed.
We follow a true end-to-end learning approach and hence avoid the problematic deconvolution.
We evaluate our approach on the data of the MR CLEAN trial and show the contributions of the various components of the method.

\subsection{Related work}
Estimation of tissue status and prediction of the final infarct based on perfusion measurements is a vast research domain.
Early work aimed mainly at finding thresholds for various perfusion parameters
 that correlated well with the final infarct in patients with recanalization (with the final infarct closely related to the core)
 and patients without recanalization (final infarct equal to the perfusion lesion) \citep{Wintermark2006, Bivard2013}.
These approaches are currently used in clinical practice.

Various works have suggested to use more advanced machine learning techniques to predict core and penumbra.
Most of them use MR imaging which usually consists of two modalities: MR Perfusion, which is similar to CTP, and diffusion weighted imaging (DWI)
  \citep{Wu2001,Scalzo2012,McKinley2017,Nielsen2018}.
The latter modality gives a clear signal regarding the viability of tissue and is considered the golden standard to identify ischemic tissue, making the prediction of final infarct easier than using only perfusion information.
However, since CTP is more prevalent in clinical practice -- even in clinical trials, it is two to three times more used than MR imaging \citep{Albers2018} -- and has a larger need for improved processing, we find it valuable to focus on this modality. 

With the advent of mechanical thrombectomy, accurate determination of the time of recanalization became possible and this parameter was introduced in models.
\citet{DEsterre2015} investigated the influence of the time between onset and imaging and the time between imaging and recanalization,
 finding that the former does not and the latter does influence the optimal perfusion parameters thresholds to determine the final infarct.
\citet{Kemmling2015} proposed a multivariate generalized linear model that predicts the final infarct in a voxelwise fashion 
 based on CTP perfusion parameters and clinical parameters including a binary recanalization status and the onset-to-recanalization duration.
It was trained and cross-validated on 161 subjects that underwent CTP imaging and mechanical thrombectomy. 
In a similar vein, the ISLES 2016 \& 2017 challenges \citep{Winzeck2018}
 invited researchers to evaluate methods that predict the final infarct based on acute DWI and MR Perfusion imaging.
The organizers provided publicly available training data and a blinded test set for the validation.
The challenge showed the competitive advantage of deep neural networks.
Most methods use a voxelwise prediction to estimate the lesion, but other approaches are possible: \citet{Lucas2018} demonstrate on a dataset of 29 subjects a novel method that predicts the core and the perfusion lesion in a voxelwise fashion, but predicts the final infarct by interpolating between the former segmentations in a learnt shape space. This allows incorporating specific domain knowledge in the training process -- which improves the results in their validation -- but also requires several assumptions -- which might limit the performance if larger training sets are used.

All approaches discussed so far depend on deconvolution as a first step, but there has been some work to avoid explicit deconvolution.
On one hand we can distinguish alternative perfusion parameters that do not require deconvolution, such as the first moment of the time-concentration curve \citep{Christensen2009}.
On the other hand, we see approaches that aim at replacing the traditional deconvolution analysis with a learned alternative \citep{Ho2016, Hess2018}.
However, this approach seems to beg the question, since during training the ground truth perfusion parameters are provided by deconvolution analysis.
In our approach -- starting from the native images and optimizing the network for the best final infarct prediction -- the optimal features and how to calculate them is inferred directly from the data.
The disadvantage of this approach is that those features no longer have a clear physiological interpretation such as the perfusion parameters have,
 but depending on the application, this might be a price worth paying.
Recently \citet{Pinto2018} proposed a method for the ISLES 2017 challenge that uses, apart from the DWI image and the perfusion parameters, also the native MR perfusion measurements to predict the final infarct.
They show that their method benefits from the native measurements, but the obtained results are not as good as other state-of-the-art approaches that did not include native measurements,
 making it hard to assess the added value of the native measurements.
The method did not take the reperfusion status or other treatment data into account,
 presumably because the dataset was fairly homogenous, consisting of mostly early reperfused patients.
 \citet{Yu2017} predict hemorrhagic transformation in a voxelwise fashion from DWI and MR Perfusion imaging and show that using both native modalities outperforms using only the permeability parameter.

\section{Methods}
We propose to use a voxelwise classification approach, where a neural network learns the relation between the final infarct status of a voxel (the output) and the CTP measurements and the metadata (the input).
The CTP measurements consists of the time-attenuation curves of that voxel, its neighboring voxels and the voxel of the AIF.
The metadata consists of four values:
 the time between stroke onset and CTP imaging,
 the time between CTP imaging and the end of the mechanical thrombectomy,
 the mTICI scale after thrombectomy
 and the persistence of the occlusion at 24 hours CT angiography.
 
\subsection{Preprocessing}
There is considerable variability in the acquisition protocol of CTP images and, due to the long acquisition time, the patient might move.
During preprocessing, we account for both aspects, aiming to reduce the unnecessary variation that the network need to cope with.
First, if the CTP was acquired with gantry tilt, the CTP images are resampled to an orthogonal grid that has the same x-y plane and an orthogonalized z-axis.
Second, the CTP scan is motion corrected by rigidly aligning the volumes with the first volume.
Registration is performed using Elastix, optimizing for the sum of squared differences.
Finally, the images are spatially downsampled to 1.5 x 1.5 x 4 mm$^3$ and temporally resampled to one image per two seconds.
The rationale is that the original high spatial resolution might help the registrations, whereas for the perfusion analysis,
 which has a notoriously low Signal to Noise Ratio (SNR), a high resolution would only slow down processing.
If the CTP was acquired in shuttle mode, i.e. with the patient continuously moving back and forth in the scanner resulting in variable scan times per CTP volume, 
 we account for this and resample the time series such that each volume corresponds to a single time point.

The creation of the ground truth -- i.e. the final infarct status of each voxel in the CTP -- and the selection of the arterial input function is described later in Sec.~\ref{sec:dataset}. 

\subsection{Network}
\label{sec:network}
 
\begin{figure*}
	\centering
	\includegraphics[width=.9\textwidth]{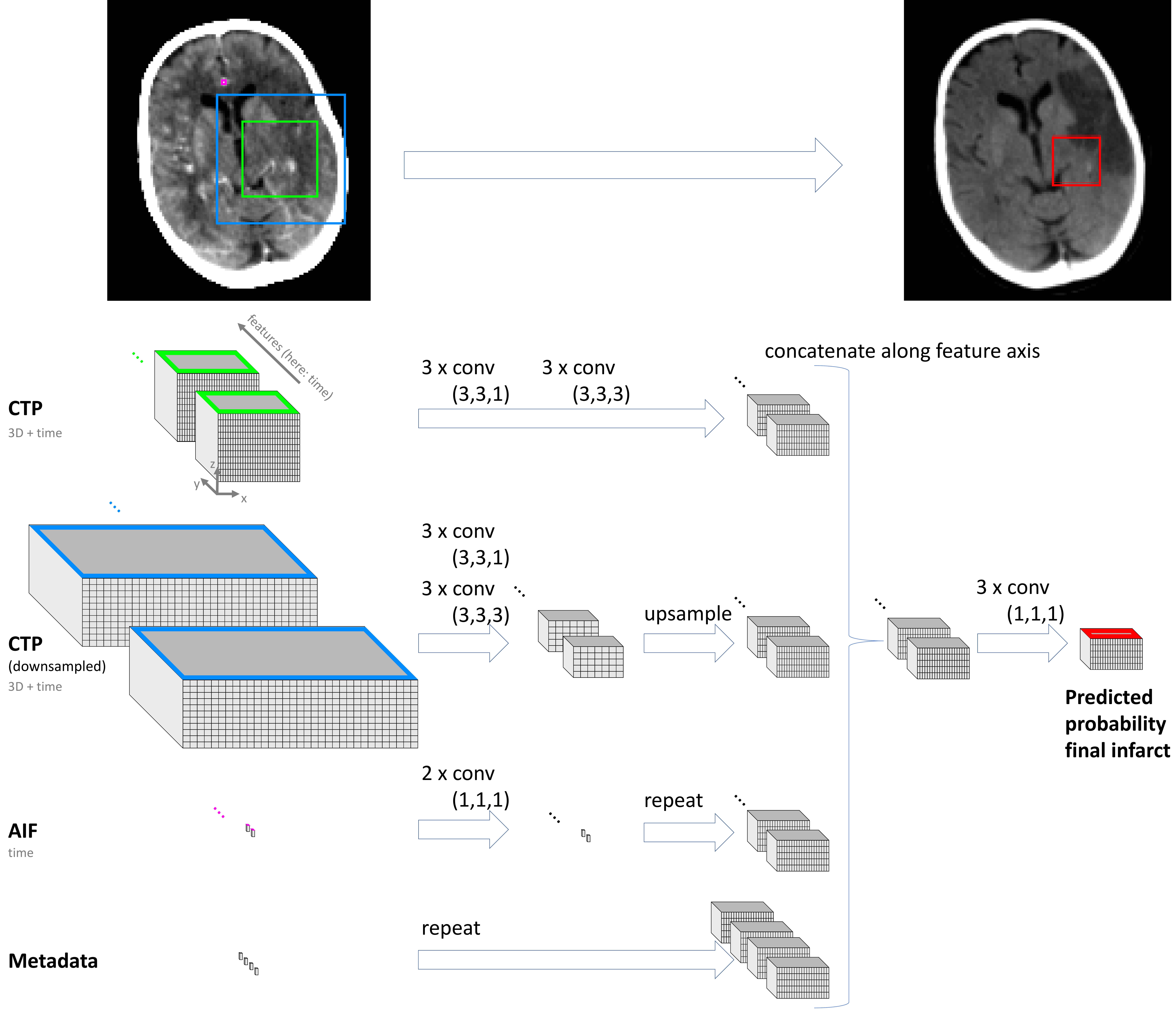}
	\caption{The proposed neural network. There are four inputs: the CTP (3D+time) the downsampled CTP (3D+time), the arterial input function (AIF, time) and the metadata (1D). The network has a fully convolutional structure, where each input can go through a series of convolutional layers before being upsampled to the original resolution. The outputs of the four pathways are concatenated and are passed through 3 convolutional layers with a 1x1x1 kernel (acting as fully connected layers) before the prediction is given. }
	\label{fig:network}
\end{figure*}

Our network is an extension of the model proposed by \citet{Kamnitsas2017} and has four different inputs.
Each input is followed by series of operations such as convolutions and upsampling, called the pathway,
 and finally the outputs of the pathways are concatenated and fed into a common pathway that gives the voxelwise prediction.
The first input is the CTP image which has 3 spatial dimensions plus time.
The input data is anisotropic, with a higher resolution in-plane than out-of-plane. However, from a clinical perspective, there is no need to cover more distance in the z-direction than in the other directions.
Hence, we compensate for this by using anisotropic kernels:
 the first pathway consists of 6 convolutional layers: three layers with 48 filters of 3x3x1 and three layers with 64 filters of 3x3x3.
The second input is the downsampled CTP. 
The image is downsampled with a factor 3 in-plane, resulting in voxels of 4.5 x 4.5 x 4 mm$^3$ and which provides the network a wider spatial context to base its prediction on.
After downsampling, the voxels are almost isotropic, but the field of view of the CTP scans is typically not: in-plane the complete brain is covered, but along the z-axis the brain coverage is usually limited (see also Sec.~\ref{sec:brain_coverage}). Therefore we decided to use anisotropic kernels to make the receptive field of the output neurons larger in-plane.
This second input's pathway has the same architecture as the first one and is followed by an upsampling operation to recover the original resolution.
The first and second pathway do not share their weights.
The third input is the arterial input function (AIF), which is a one dimensional vector. 
This pathway consists of 2 convolution layers, each with 48 filters of 1x1x1, and an upsampling operation.
The last input is the metadata (1D) and this pathway only consists of an upsampling operation.
At this point, the outputs of the four pathways have the same spatial dimensions and are concatenated in the feature dimension.
This resulting output is fed through 3 convolutional layers with each a 1x1x1 kernel. The first two have 150 filters and the final one has 1 filter and gives the prediction.

In all convolutional layers, we use batch normalization \citep{Ioffe2015} and the PreLU activation function \citep{He2015}.
In total, this network has 774,787 trainable parameters.
 
This network has a fully convolutional structure, which allows to use dense inference and training.
By predicting multiple adjacent voxels simultaneously,
 the computational efficiency is increased because redundant calculation in the overlapping receptive fields of the output neurons are avoided.
We use an output size of 21x21x5, giving the architecture visualized in Fig.~\ref{fig:network}.

The input intensities are normalized by first clipping them to the range of -100 to 1000 HU and then linearly transforming them to have mean of zero and a standard deviation of one.

During training, the network is optimized by stochastic gradient descent with Nesterov momentum for the weighted cross-entropy,
 with the positive class weighted ten times as heavy, resulting in balanced training.
The training samples are uniformly chosen from within the intracranial volume and augmented by flipping samples left/right, small rotations, Gaussian noise
 and a CTP specific augmentation we introduced in \citep{Robben2018SWITCH}.
 
The CTP specific data augmentation exploits the fact that the perfusion measurements are a linear time invariant system.
This means that, if the contrast injection was a bit later, both the AIF and the CTP timeseries would show the same delay.
Similarly, if the injection was earlier, all curves would shift to the left.
However, this has no influence on the actual tissue perfusion status or the viability of the tissue.
If the concentration of the iodine in the contrast agent were a fraction higher or lower, the concentration curves would change with the same fraction.
But again, this has no impact on the tissue status.
Hence, we augment our training dataset by applying a random time shift (earlier or later) and a random scaling of the attenuation variation.
The shift is randomly chosen between -4 and 6 time points (since our measurements are discrete)
 and the scaling has a log normal distribution between with $\mu$ of 0 and $\sigma$ of 0.3.

The method is implemented in Python using Keras and DeepVoxNet \citep{Robben2018DeepVoxNet}.

\subsection{Ablation study}
We want to measure the impact of the different design choices
 and hence we gradually ablate our method to understand how the different components contribute to the performance.

\subsubsection*{The impact of no deconvolution}
To quantify the impact of the deconvolution-free approach we benchmark a network working on the deconvolved timeseries.
To obtain the deconvolved time series, we use Tikhonov regularized SVD-based deconvolution \citep{Fieselmann2011}
 with the Volterra discretization scheme \citep{Sourbron2007} -- to which we simply refer as deconvolution.
The deconvolution has one hyperparameter, the relative regularization parameter $\lambda_{rel}$ which determines the amount of filtering and which is set to 0.4.
As the deconvolution is very noise sensitive, we first perform a spatial Gaussian smoothing with an isotropic sigma of 2.5\,mm.
To make the comparison fair, we also test the proposed (i.e. the non-deconvolution based) network on this smoothed data - called \textbf{Proposed (smoothed)}.
The deconvolved time series are used as input to the network \textbf{Proposed (deconvolved)}.
This network and its training is identical to the one described in Section \ref{sec:network}, with the exception that the AIF is no longer provided and the CTP specific data augmentation is turned off.

\subsubsection*{The impact of spatial context}
Current clinical systems do not take the spatial context into account to determine the tissue status of a voxel.
We are interested in to what extent the spatial context affects the predictions.
Hence we introduce two extra networks: \textbf{One-voxel (smoothed)} and \textbf{One-voxel (deconvolved)}.
Both only have three 1x1x1 convolutions on the original resolution and do not use the subsampled version and hence do not use spatial context.
Otherwise, both are identical to the earlier introduced methods.

\subsubsection*{The impact of the CTP data augmentation}
We investigate the impact of our CTP specific data augmentation and train the proposed network also without the augmentation.

\subsubsection*{The impact of the metadata}
Our network not only uses the imaging data, but also the treatment parameters.
In current clinical practice, there is only a distinction between core and penumbra -- roughly corresponding to immediate successful thrombectomy and no treatment at all.
The additional treatment parameters our method uses might improve the accuracy of the predictions.
To that end, we evaluate three variants of the proposed network. 
One that uses binarized mTICI scores (0-2a vs 2b and 3), one that does not include time from onset to CTP and one that does not have time from CTP to end of thrombectomy.

\subsubsection*{The impact of multi-scale input}
Our proposed network works on two scales on the CTP images.
\citet{Kamnitsas2017} showed the benefit of that approach for various segmentation tasks.
We perform two ablation experiments, one where we remove the high-resolution pathway and one where we remove the low-resolution pathway, to see if those conclusions also hold for our application. Both these networks have 405,283 parameters.

\section{Experiments}
We evaluate the proposed method and its variants on the data of the MR CLEAN trial (described in more detail in the next section, Sec.~\ref{sec:dataset}).
In all experiments, we do a five-fold cross-validation, and report the results aggregated over all the left-out subjects.
The hyperparameters of the optimization are experimentally set based on preliminary experiments on the first fold.

\subsection{Dataset}
\label{sec:dataset}
\begin{figure}
	\centering
	\includegraphics[width=.4\textwidth]{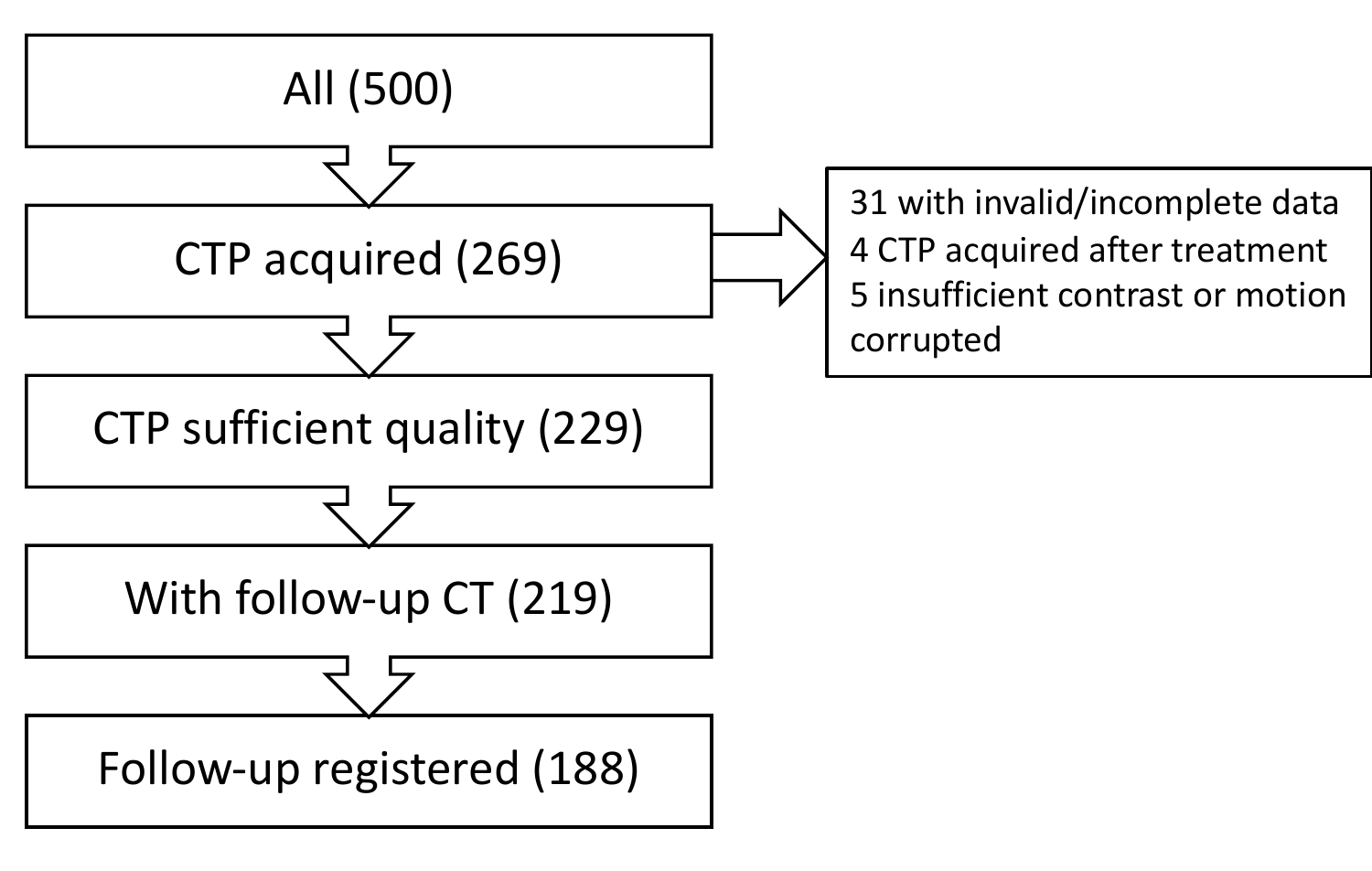}
	\caption{Subject selection. }
	\label{fig:subjects}
\end{figure}

MR CLEAN was a multicenter study to investigate the benefit of thrombectomy in acute ischemic stroke \citep{Berkhemer2015}.
The study's inclusion criteria are described in detail in the study protocol;
 in short: patients were randomized within 6 hours after stroke onset and when a large vessel occlusion was identified in the anterior circulation. 
In our work, we include all MR CLEAN subjects that had a baseline CTP of sufficient quality and a follow-up non-contrast computed tomography (NCCT) that could be registered to the acute CTP.
This results in 188 included subjects. The selection process is shown in Fig.~\ref{fig:subjects}.
The patient characteristics are shown in Table~\ref{tab:characteristics}.

\begin{table}[]
	\centering
	\begin{tabularx}{0.7\textwidth}{ l r }
	Mean (IQR) age [years] 			&	62 (52 - 73)\\
	Men [\%] 									&	55\\
	Treatment with IV tPA [\%]		&	87\\
	Intra-arterial treatment [\%]		&	45\\
	Location of occlusion on CTA [\%]\\
	\quad    ICA 								& 27\\
	\quad    M1	 							&	67\\
	\quad   M2		 						&	5\\
	\quad    A1									&	0\\
	\quad    A2								&	1\\
	Mean (IQR) duration from \\
	\quad stroke onset to imaging [min] 		&	174 (108-239) \\
	\quad imaging to end IAT [min] 				&	180 (138-214)\\
	Mean (IQR) final infarct volume [ml] 	 	& 78 (21 - 121)\\
    Recanalisation in IAT (mTICI 2B/3) [\%] &	67\\
	\end{tabularx}
	\caption{Characteristics of the 188 included subjects from the MR CLEAN dataset.  }
	\label{tab:characteristics}
\end{table}

The MR CLEAN study protocol prescribes follow-up imaging with NCCT at 24 hours and 5 days after onset.
Where available, we use the 5 day follow-up otherwise the 24 hour follow-up is used.
The follow-up images were semi-automatically delineated and reviewed by an experienced reader \citep{Boers2013,Bucker2017}.

The follow-up NCCT images are registered to the acute CTP images.
First a rigid registration between the follow-up and the acute CTP is performed.
Afterwards, a non-rigid registration is done in an attempt to compensate for brain swelling due to edema.
The found transformations are applied to the final infarct delineations.
Finally, we segment the cerebrospinal fluid (CSF) -- in particular the ventricles -- from the acute CTP images,
 and exclude those voxels from the transformed final infarct,
 under the reasoning that CSF cannot infarct and any overlap between CSF and final infarct is rather due to misalignment.

We manually select in each CTP image an arterial input function (AIF) and use those in all experiments.
Hereto, we select the time series of the voxel that shows the largest increase in attenuation coefficient and is located in an unaffected proximal artery.

\subsection{Prediction of the final infarct}
The prediction of the final infarct is the most important capability of the network.
For current clinical practice, the most important metric is the volume of the final infarct,
 and hence we focus on this metric and the difference in volume between the predicted volume and true volume.
As predicted volume, we take the sum of all predicted probabilities, multiplied with the volume per voxel.

From a scientific point of view, the localization of the infarct is also interesting, and hence we also report the Dice and the soft Dice scores.
The latter is like the standard Dice score, except that the predictions are not binarized.
Non-parametric paired significance tests (bootstrapping) are used to test the significance of the results.

\begin{figure}
	\centering
	\includegraphics[width=.45\textwidth]{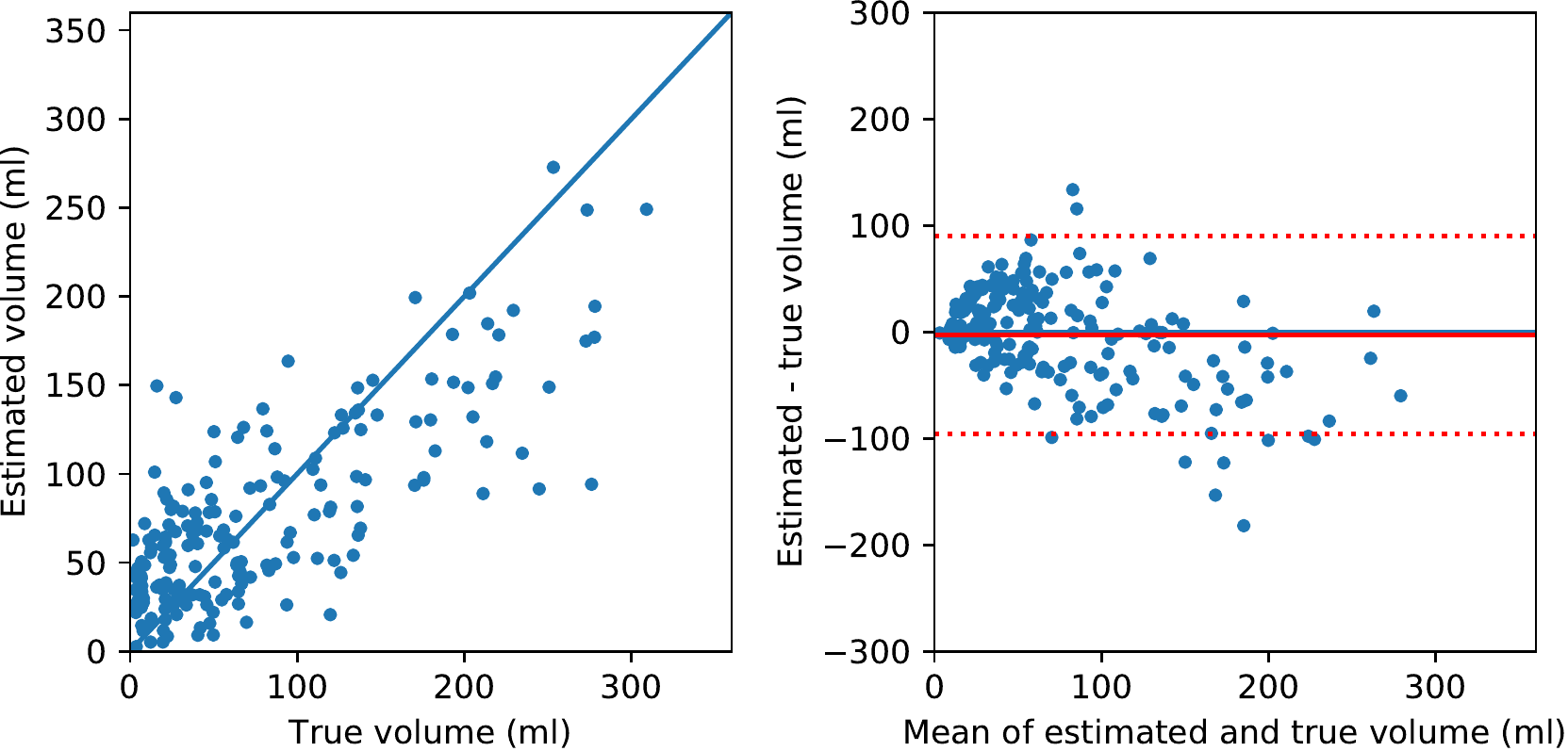}
	\caption{The true and predicted final infarct volumes for all subjects in the testing folds, using the proposed method. }
	\label{fig:prediction_scatter}
\end{figure}

\begin{figure*}[tb]
	\centering
	\subfloat[Subjects with the 20\%, 40\%, 60\% and 80\% percentile Dice score in the subgroup with a true final infarct volume less than 100 ml. ]{
		\includegraphics[width=4.5cm]{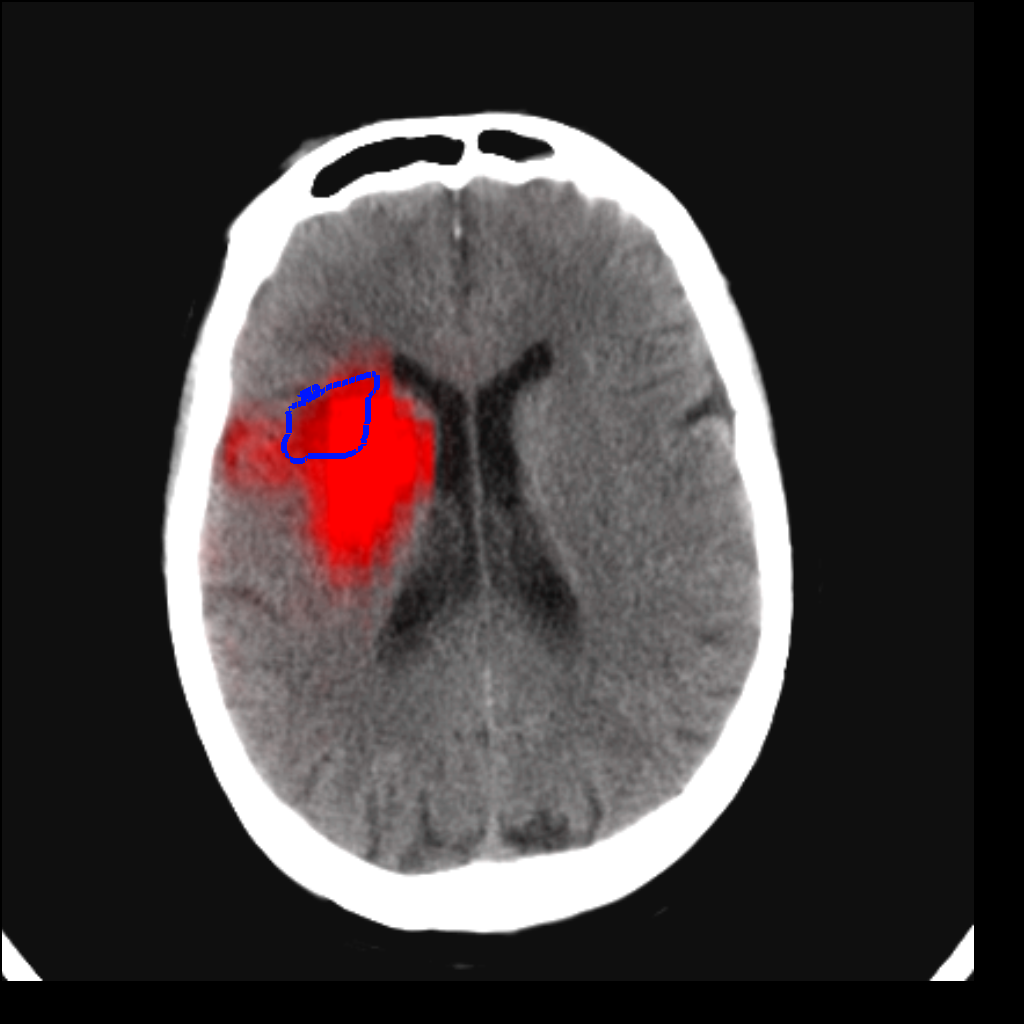} 
		\includegraphics[width=4.5cm]{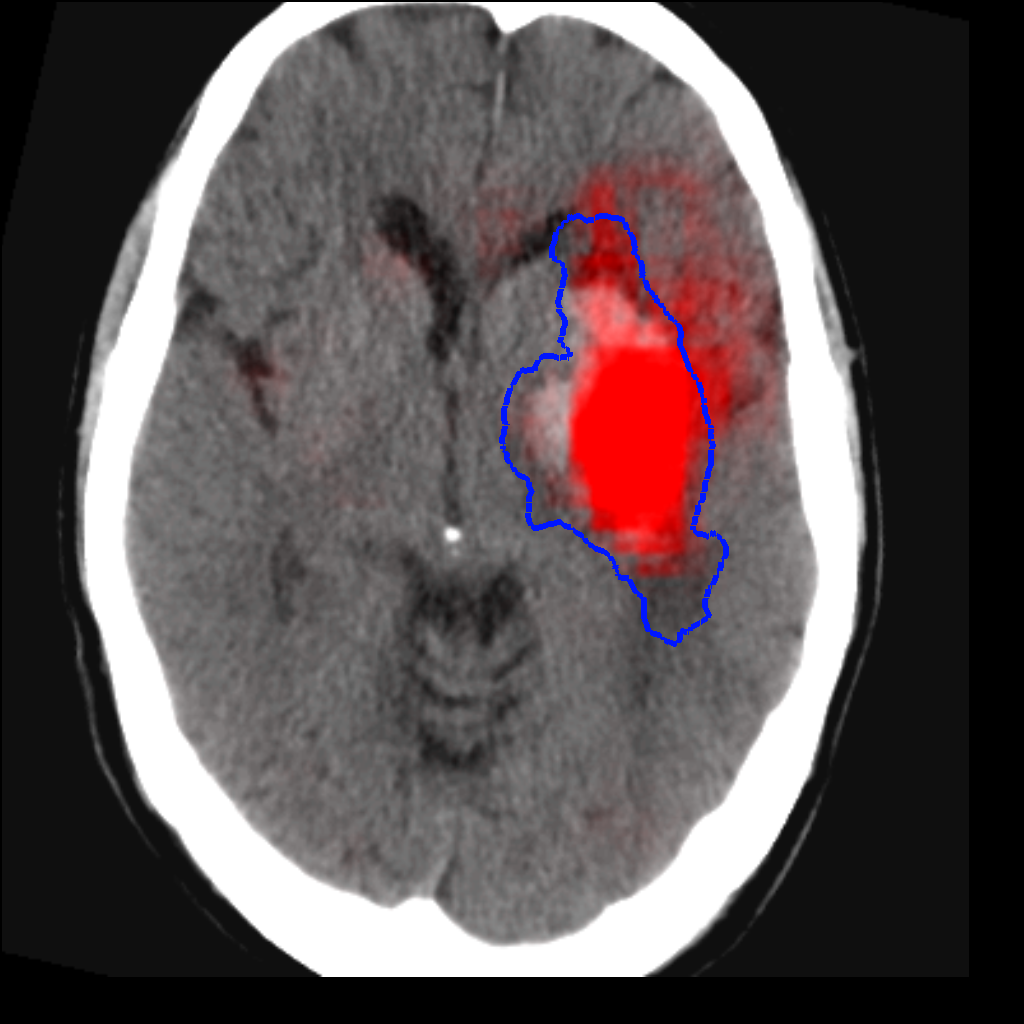}
		\includegraphics[width=4.5cm]{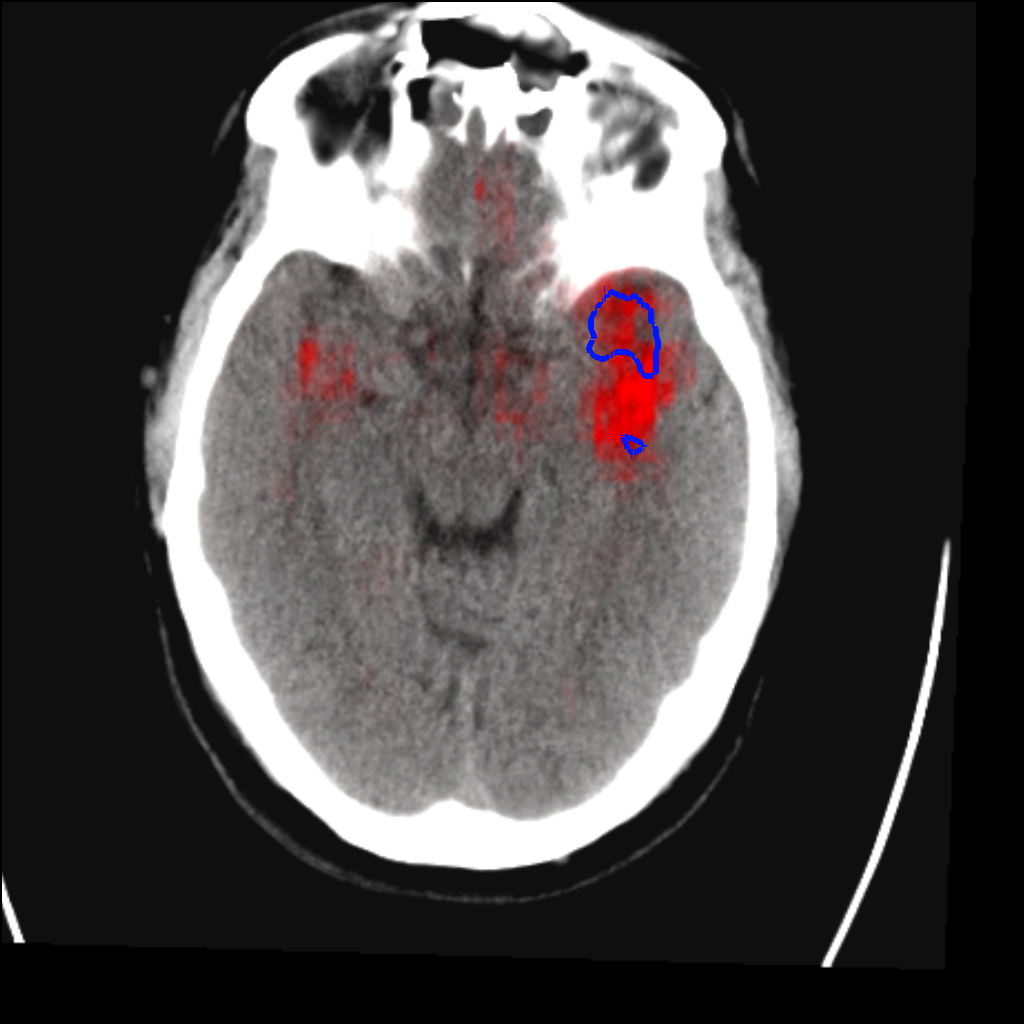} 
		\includegraphics[width=4.5cm]{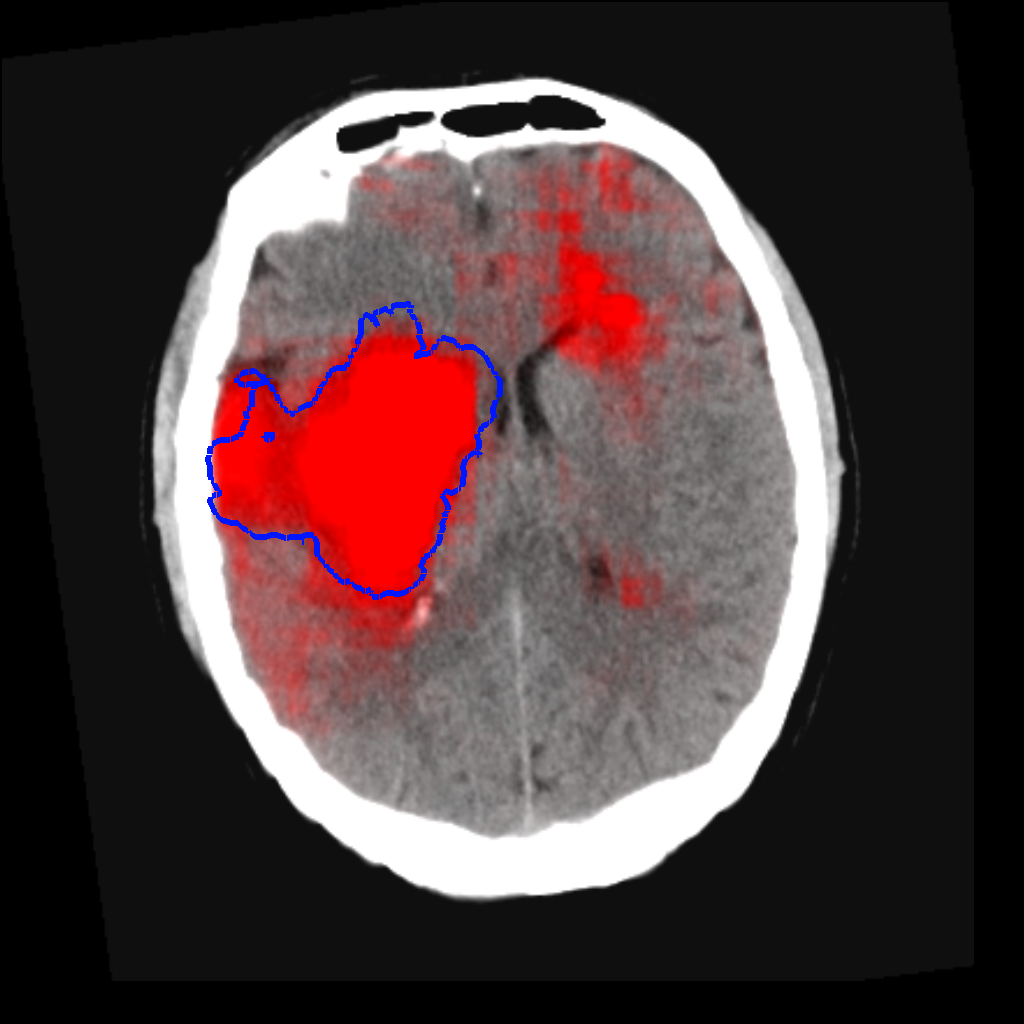}
		\label{fig:examples_small}
	}
	
	\subfloat[Subjects with the 20\%, 40\%, 60\% and 80\% percentile Dice score in the subgroup with a true final infarct volume of at least 100 ml. ]{
		\includegraphics[width=4.5cm]{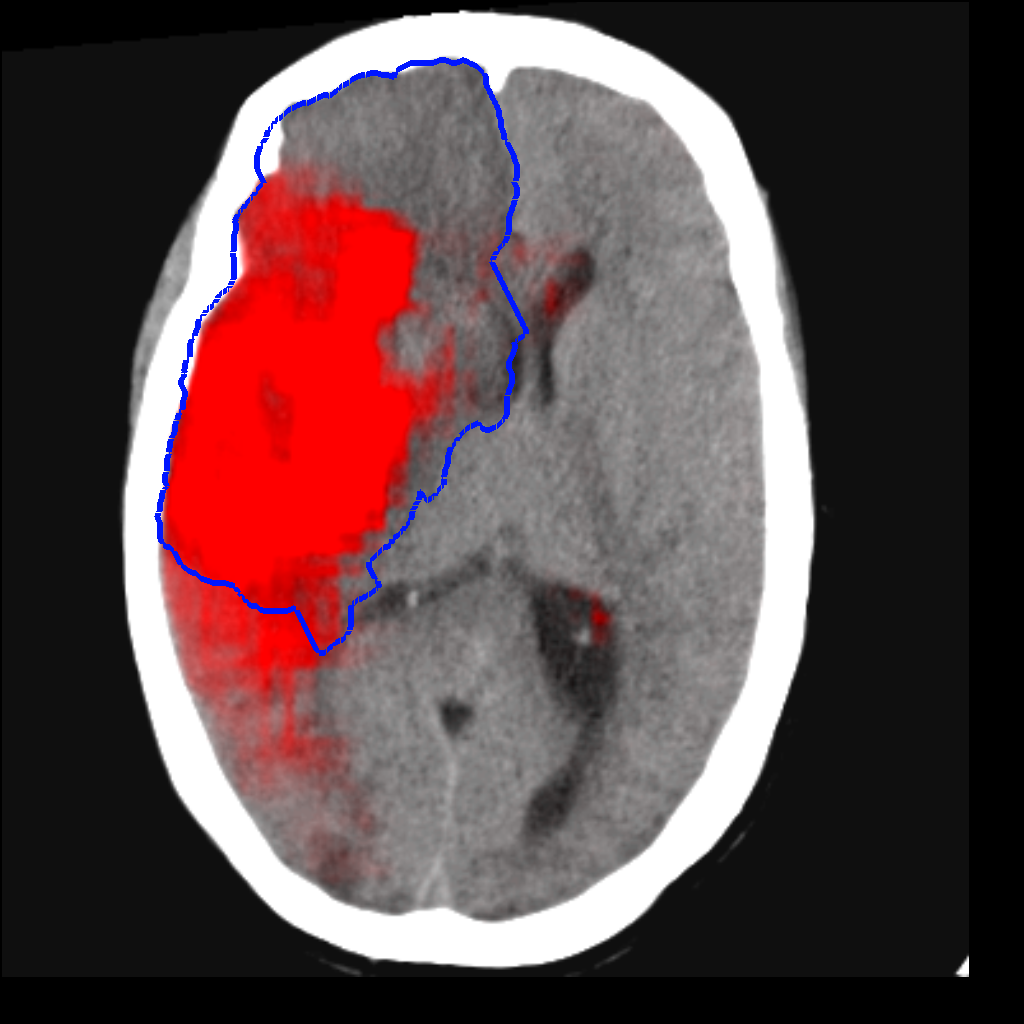} 
		\includegraphics[width=4.5cm]{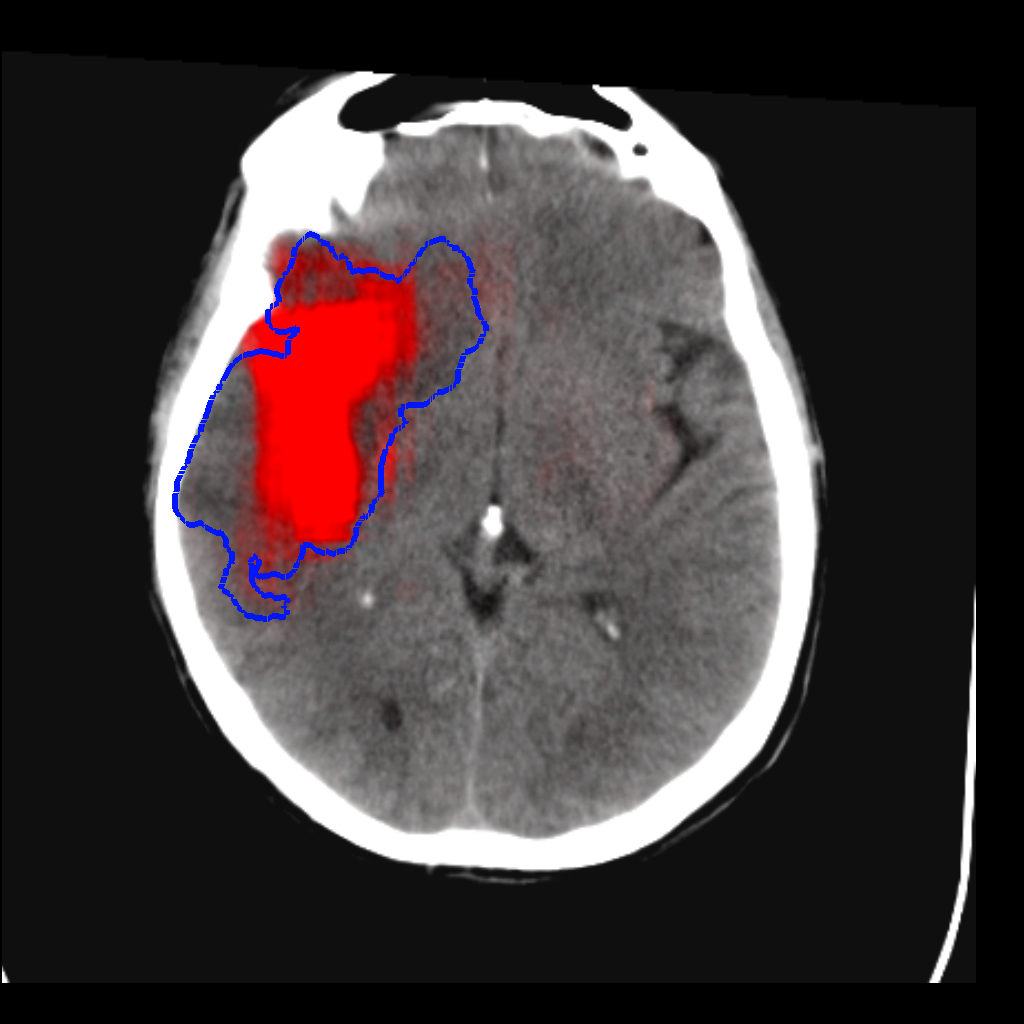}
		\includegraphics[width=4.5cm]{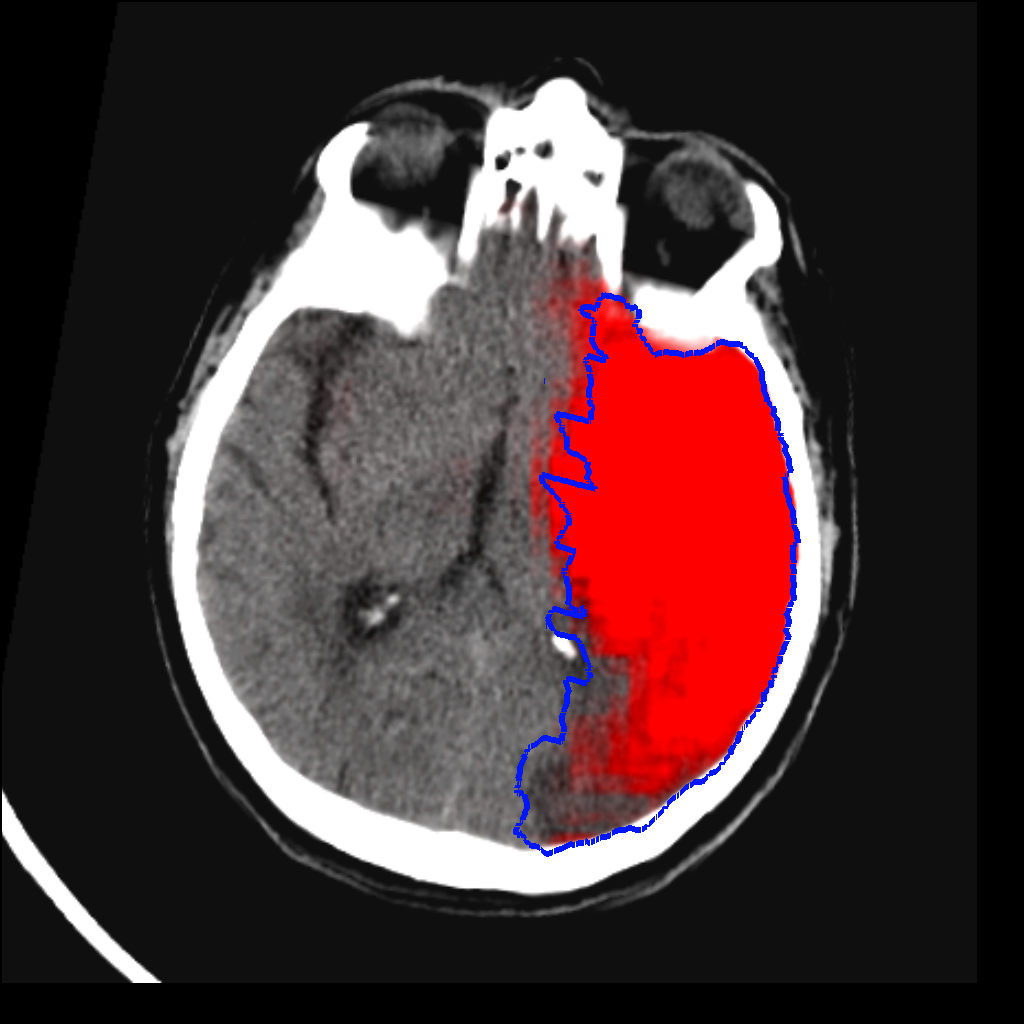} 
		\includegraphics[width=4.5cm]{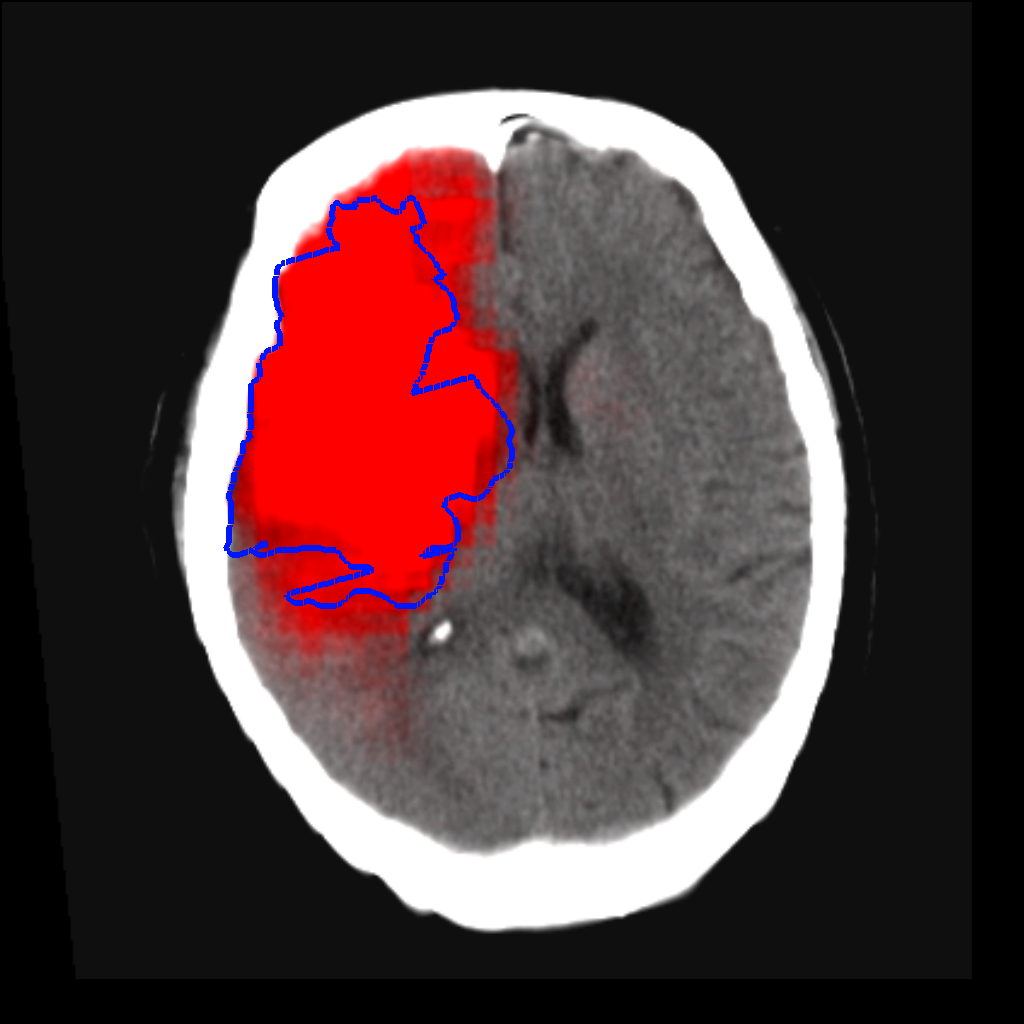}
		\label{fig:examples_large}
	}
	\caption{Predictions of the proposed method on a representative sample of subjects from the test set. The probabilistic predictions are overlayed in red on the follow-up NCCT scan whereas the ground truth final infarct is outlined in blue. }
	\label{fig:examples}
\end{figure*}

\newcolumntype{C}{>{\centering\arraybackslash}X}
\newcolumntype{L}{>{\raggedright\arraybackslash}X}
\newcolumntype{R}{>{\raggedleft\arraybackslash}X}

\begin{table*}[]
	\centering
	\begin{tabularx}{\textwidth}{
	 >{\hsize=.05\hsize}L
	 >{\hsize=1\hsize}L
	 >{\hsize=.2\hsize}R
	 >{\hsize=.2\hsize}L
	 >{\hsize=.2\hsize}R
	 >{\hsize=.2\hsize}L
	 >{\hsize=.2\hsize}R
	 >{\hsize=.1\hsize}L
	 >{\hsize=.25\hsize}L
	 >{\hsize=.2\hsize}R
	 >{\hsize=.2\hsize}L
	 }
	 \multicolumn{2}{l}{\multirow{2}{*}{Method} }
		&  \multicolumn{2}{c}{Mean}			& \multicolumn{2}{c}{Mean}
		& 	\multicolumn{3}{c}{Mean absolute}		& 	\multicolumn{2}{c}{AUC}
	 \\
	&	&  \multicolumn{2}{c}{ soft Dice} 	& \multicolumn{2}{c}{Dice}
		&  \multicolumn{3}{c}{volume error (ml)}	& 	\multicolumn{2}{c}{precision-recall}\\
	\hline
	\hline
	\multicolumn{2}{l}{Proposed} & \textbf{0.40} &  & \textbf{0.48} &  & \textbf{36.7} &  & & \textbf{0.54} & \\ 
	\hline
	\multicolumn{2}{l}{Proposed (smoothed)} & 0.39 & (**) & 0.45 & (**) & 37.5 &  & + 2 \% & 0.53 &  \\ 
	\multicolumn{2}{l}{Proposed (deconvolved)} & 0.36 & (**) & 0.42 & (**) & 39.1 &  & + 6 \% & 0.50 & (**) \\ 
	\multicolumn{2}{l}{One-voxel (smoothed)} & 0.21 & (**) & 0.15 & (**) & 45.4 & (**) & + 23 \% & 0.37 & (**) \\ 
	\multicolumn{2}{l}{One-voxel (deconvolved)} & 0.17 & (**) & 0.04 & (**) & 48.6 & (**) & + 32 \% & 0.34 & (**) \\ 
	\hline
	\multirow{2}{*}{\rotatebox{90}{\parbox{3.0cm}{\centering Ablation proposed}}} 
	& no data augmentation & 0.39 & (*) & 0.46 & (**) & 41.5 & (**) & + 13 \% & 0.50 & (**) \\ 
	\cline{2-11}
	& binary mTICI & 0.40 &  & 0.48 &  & 38.5 & (*) & + 5 \% & 0.53 &  \\ 
	& no time from onset to CTP & 0.40 &  & 0.48 &  & 38.4 & (*) & + 5 \% & 0.54 &  \\ 
	& no time from CTP to end thrombectomy & 0.39 & (**) & 0.47 & (*) & 41.1 & (**) & + 12 \% & 0.54 &  \\ 
	\cline{2-11}
	& no hi-res pathway & 0.38 & (**) & 0.44 & (**) & 37.6 &  & + 2 \% & 0.50 & (**) \\ 
	& no lo-res pathway & 0.35 & (**) & 0.41 & (**) & 38.4 &  & + 4 \% & 0.52 & (*) \\ 
	\end{tabularx}
	\caption{Error metrics between the predicted and ground truth final infarct. A paired significance test is performed between the proposed method and its variants, with (*) indicating $P<0.05$ and (**) indicating $P<0.005$.  }
	\label{tab:results}
\end{table*}

\subsection{Prediction of core and perfusion lesion}
It is interesting to see how the metadata influences the prediction of the network,
 in particular the difference between immediate recanalization and no recanalization. 
To this end, we predict for each subject in the test set the hypothetical final infarct volume 
in the case of early complete recanalization (mTICI 3 at 60 minutes) and in case of no recanalization at all.
In the former scenario, the predicted final infarct should correspond to the infarct core at baseline imaging,
 whereas in the latter scenario, it should correspond to the perfusion lesion.
 
\begin{figure}
	\centering
	\includegraphics[width=.45\textwidth]{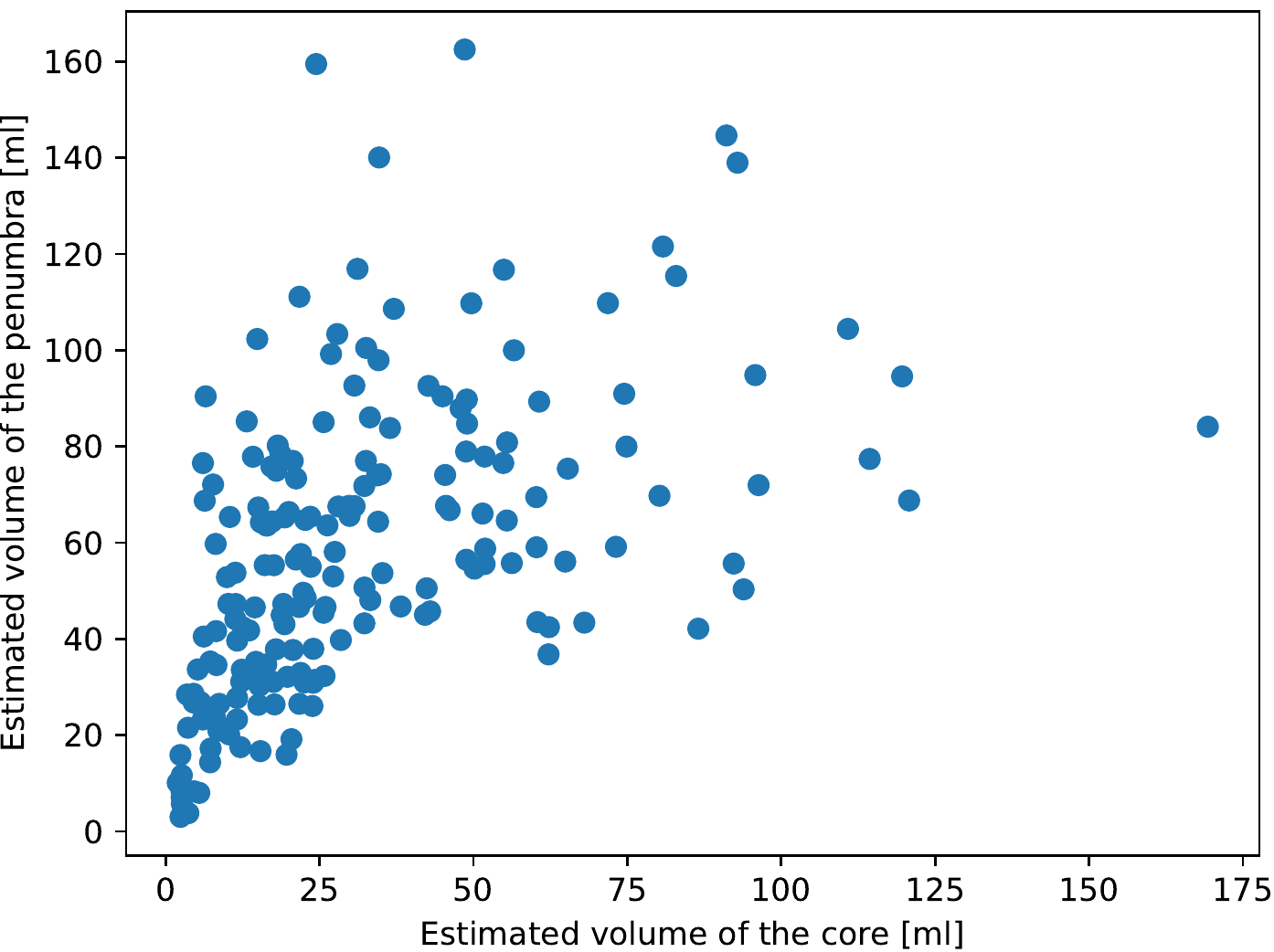}
	\caption{The predicted core and penumbra volumes for all subjects in the testing folds, using the proposed method. }
	\label{fig:core_penumbra_volumes}
\end{figure}

\subsection{Influence of the AIF selection}
The prediction of the network depends on the AIF.
To investigate the sensitivity of the method on the AIF selection,
 we randomly select 15 subjects and reannotate the AIF according to the same guidelines, with more than 6 months between the first and second annotation.
We predict the final infarct volume for those subject with both AIFs and compare the two predictions.

\begin{figure}
	\centering
	\includegraphics[width=.4\textwidth]{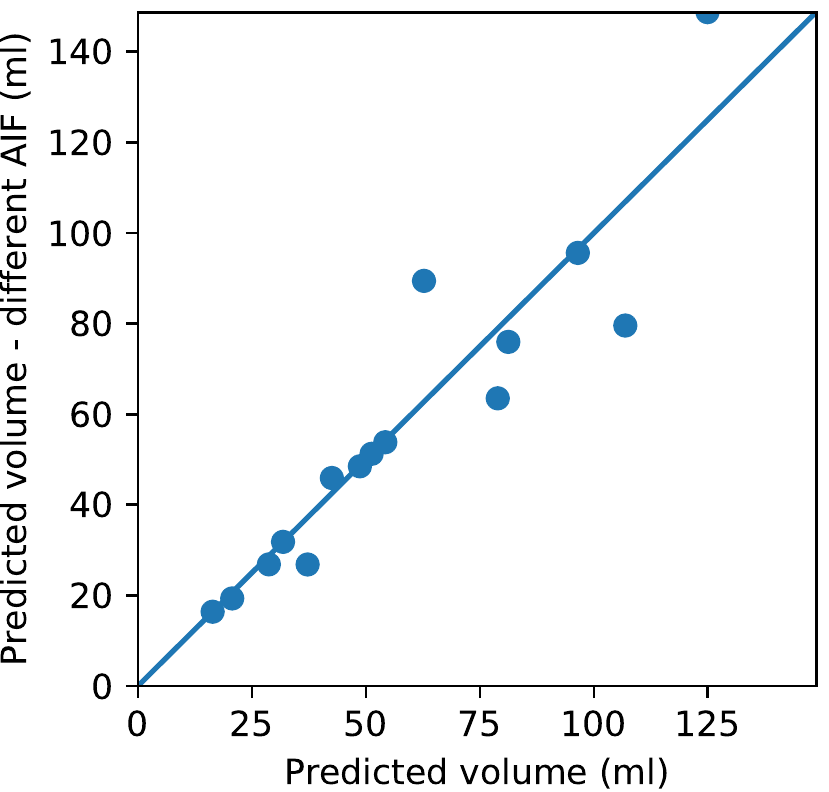}
	\caption{The predicted final infarct volume of 15 randomly selected subjects with the AIF annotated twice by the same obsever, the second time more than six months later, illustrating the limited influence of the manual AIF selection on the final infarct prediction.  }
	\label{fig:robustness_AIF}
\end{figure}

\section{Results and discussion}
Fig.~\ref{fig:prediction_scatter} shows a scatter plot and a Bland-Altman plot of the true and predicted volumes for all subjects in the testing folds using the proposed method. The mean volume error is -2.8 ml (the prediction is a slight underestimation) and the mean absolute volume error is 36.7 ml. The mean Dice score is 0.48. 
A representative set of predictions is shown in Fig. \ref{fig:examples}.

The achieved Dice and volume errors seem respectively very low and high.
We believe nevertheless that these are state-of-the-art results on an inherently difficult problem.
This is not an image segmentation task, where all the relevant information is contained in the input,
 but a prediction problem with a ground truth that is defined on images acquired several days later.
For example, the ISLES 2017 challenge \citep{Winzeck2018} has a similar goal,
 namely predicting the final infarct from acute DWI and MR perfusion imaging,
 and reported a mean Dice score of $0.32$ the best result.
Of course, a direct comparison between these values is not possible since the ISLES challenge had different modalities (with especially the DWI imaging being very informative compared to perfusion imaging) and a different population (mostly early successful early recanalization, which results in small lesions and hence lower Dice scores),
 but it illustrates the difficulty of the problem.

\subsection{The impact of spatial context, no deconvolution and data augmentation}
Table~\ref{tab:results} shows the mean Dice, soft Dice and absolute volume error of the proposed method and of the various ablated variants.
It shows a gradual decrease in performance, both for removing the spatial context as for working on the deconvolved data.
The beneficial effect of taking context into account is expected.
Perfusion imaging is a noisy modality and suffers from artifacts which results in -- often visually -- erroneous predictions.
In some clinical studies, the results given by the conventional voxelwise perfusion analysis are manually corrected.
We hypothesize that a human rater is able to do so, because she takes spatial context into account.
By providing spatial context, the network is able to do this as well, which greatly improves results: 
 the mean Dice score more than triples and the volume error drops almost 20\%.
Our network uses both high and low resolution versions of the CTP image simultaneously.
The ablation experiment shows that both are beneficial, with the low resolution a bit more important (i.e. the ablation experiment without the low resolution pathway performs worst).

The benefit of working on the native data is more remarkable.
It can be argued that, if we have a correct physical model, there is no need to re-learn that model a second time from our own limited training data.
From the perspective of physics, deconvolution is the right way to approach the problem.
However, the way that deconvolution should be regularized is not dictated by physics.
We hypothesize that a neural network can do deconvolution better as it has a learned model of the noise and concentration curve shape.
We showed this earlier on simulated data \citep{Robben2018SWITCH} and see this now confirmed on real data:
 using the native data instead of the deconvolved data improved the mean Dice score significantly from 0.42 to 0.48.
Note however that learning this deconvolution implicitly is data-intensive and the proposed data augmentation is a necessary part, even with our relatively large training set.

Apart from the improved performance, a second advantage of avoiding the explicit deconvolution, is that the selection of the AIF can become part of the network.
In this work, we used manually selected AIFs for all experiments, but AIF selection is a difficult problem and has great influence on the deconvolution results.
However, if the AIF is an input to the network, it becomes possible to let the network learn to select the AIF from an input patch
 and have the AIF selection optimized in an end-to-end fashion, such as was explored by \citet{Hess2018}.

\subsection{The impact of the metadata}
Fig.~\ref{fig:core_penumbra_volumes} shows the volumes of the predicted core and penumbra (perfusion lesion minus core).
It shows that the predicted final infarct volumes vary widely based on the treatment:
 the volume of the penumbra is the difference in final infarct volume between fast complete recanalization and no recanalization.
This is as expected and already widely reported in literature.

In the ablation experiments, we also investigated the influence of the metadata on the quality of the predictions.
Table~\ref{tab:results} shows that in all cases the model became less predictive,
 which shows that the model effectively uses the metadata.
Leaving out the time from CTP to end of treatment has the largest influence, increasing the mean absolute volume error with 12\%.
This shows that the network takes this information – and hence the growth of the lesion during this time – into account.
Binarizing the mTICI and leaving out the time from onset to CTP have smaller but still significant effect, both increasing the error with 5\%.
The importance of time from CTP to end of treatment and the precise mTICI scores were already earlier reported in literature \citep{Wheeler2015,DEsterre2015, Kemmling2015}.
The influence of time from onset to CTP is more surprising, as earlier literature reported this does not have an influence \citep{DEsterre2015}.

\subsection{Influence of the AIF selection}
Fig.~\ref{fig:robustness_AIF} shows the predicted final infarct volumes for 15 subject with two different AIFs.
The mean absolute volume difference between the predicted volumes is 7.80 ml, and the correlation is 0.93.
This shows that the influence of the manual AIF selection on the volume prediction is limited
 compared to the actual volume error of the predicted final infarcts.
Nevertheless, the error is large enough to make future research in automated AIF selection worthwhile.

\subsection{The quality of the ground truth}
\label{sec:quality}
An important limitation of this study is the quality of the ground truth. 
First, follow-up imaging is always with NCCT, which is less sensitive than MR.
Second, the follow-up NCCT scans are a mix of images acquired after 24 hours and after 5 days.
It is reported, on the MR CLEAN dataset, that the lesion still grows after 24 hours:
 \citet{Bucker2017} report that between these time points the median infarct volume increases from 42 ml to 64 ml
  and more than half of the subjects witness more 30\% relative growth.
This means that there is variability on our ground truth, purely due to the moment of follow-up acquisition.
Part of this volume increase is likely due to genuine delayed tissue death and part is due to edema, which is most pronounced at 3 to 5 days after the stroke.
Finally, the brain edema increases the volume of the final infarct, and hence might result in overestimated core and penumbra volumes
 even though we tried to correct for this by non-rigid registration and CSF exclusion.

We perform an additional analysis in the subgroup that had follow-up imaging at 24 hours.
For the final infarct prediction task, those 50 subjects have a mean volume error of -4.7\,ml,
 compared to -2.8 ml for the total population.
The mean absolute volume error is 42.3 ml,
 compared to 36.7 ml for the total population.
These differences are not statistically significant,
 something we mainly attribute to the limited number of subjects
 and to the lower power of the unpaired significance testing compared to paired testing.

\subsection{Limited brain coverage}
\label{sec:brain_coverage}
Our dataset is acquired on a variety of scanners, and not all of them have full brain coverage during CTP acquisition.
We find that the coverage along the axial direction is on average 65\,mm, with the first and third quartiles being 40 and 96\,mm.
As a consequence, our ground truth final infarct volume is also limited to this field of view, resulting in an underestimation of the final infarct volume.
The correlation between the final infarct volume within the CTP volume and the scan length is~0.16.

\section{Conclusion}
We have shown that a neural network can learn to predict the final infarct volume from acute CTP images and the treatment parameters.
This might help clinicians to evaluate the various treatment options.
We performed a series of ablation experiments,
 which tested the contribution of the various components of the method and showed the benefit of our deconvolution-free approach.

\subsection*{Acknowledgements \& disclosures}
David Robben is supported by an innovation mandate of Flanders Innovation \& Entrepreneurship (VLAIO) and is an employee of Icometrix.
Anna M.M. Boers, Henk A Marquering, Yvo B.W.E.M. Roos and Charles B.L.M. Majoie are shareholders of Nico.lab.
Lucianne L.C.M. Langezaal, Robert J. van Oostenbrugge and Paul Suetens report no disclosures.
Wim H. van Zwam reports speaker fees from Stryker and Cerenovus paid to institution.
Diederik W.J. Dippel and Aad van der Lugt report funding from the Dutch Heart Foundation, Brain Foundation Netherlands, The Netherlands Organisation for Health Research and Development, Health Holland Top Sector Life Sciences \& Health,  and unrestricted grants from AngioCare BV, Covidien/EV3®,  MEDAC Gmbh/LAMEPRO, Penumbra Inc., Top Medical/Concentric, Stryker, Stryker European Operations BV, Medtronic, Thrombolytic Science, LLC for research,  all paid to institution.
Charles B.L.M. Majoie reports grants from European Commssion, CVON/Dutch Heart Founation, TWIN foundation, Stryker (paid to institution).
Aad van der Lugt reports research grants from Siemens Healthineers, GE Healthcare and Philips Healthcare, all paid to institution.
Robin Lemmens is a senior clinical investigator of FWO Flanders.

The MR CLEAN trial was partly funded by the Dutch Heart Foundation and by unrestricted grants from AngioCare BV, Medtronic/Covidien/EV3®, MEDAC Gmbh/LAMEPRO, Penumbra Inc., Stryker®, and Top Medical/Concentric.

\section*{References}

\bibliography{references_without_url}

\end{document}